\documentclass{article}
\usepackage{graphicx} 
\usepackage{natbib}  
\usepackage{caption} 
\usepackage{subcaption} 
\usepackage{algorithm}
\usepackage{algorithmic}
\usepackage{amsmath,amsfonts}
\usepackage{booktabs}
\usepackage{siunitx}
\usepackage{multirow}
\usepackage{newfloat}
\usepackage{listings}

\title{Mode‑Aware Non‑Linear Tucker Autoencoder for Tensor-based Unsupervised Learning}

\author{Junjing Zheng, Chengliang Song, Weidong Jiang, Xinyu Zhang}
\date{}

\usepackage[left=4cm, right=4cm]{geometry} 

\begin{document}

\maketitle

\begin{abstract}
High-dimensional data, particularly in the form of high-order tensors,  presents a major challenge in self-supervised learning. While MLP-based autoencoders (AE) are commonly employed, their dependence on flattening operations exacerbates the curse of dimensionality, leading to excessively large model sizes, high computational overhead, and challenging optimization for deep structural feature capture. Although existing tensor networks alleviate computational burdens through tensor decomposition techniques, most exhibit limited capability in learning non-linear relationships. To overcome these limitations, we introduce the Mode-Aware Non-linear Tucker Autoencoder (MA-NTAE). MA-NTAE generalized classical Tucker decomposition to a non-linear framework and employs a Pick-and-Unfold strategy,  facilitating flexible per-mode encoding of high-order tensors via recursive unfold-encode-fold operations, effectively integrating tensor structural priors. Notably, MA-NTAE exhibits linear growth in computational complexity with tensor order and proportional growth with mode dimensions. Extensive experiments demonstrate MA-NTAE’s performance advantages over standard AE and current tensor networks in compression and clustering tasks, which become increasingly pronounced for higher-order, higher-dimensional tensors.
\end{abstract}

\section{Introduction}

High-order tensors (multi-way arrays indexed by multiple coordinates) serve as the fundamental representation for modern data-intensive applications across scientific and industrial domains \cite{Hyperspectral_eg}. Multi-view images \cite{lou_parameter-free_2025}, hyperspectral data \cite{xu_nonlocal_2019}, and spatio-temporal signals \cite{gong_esprit-based_2023} \emph{etc.}, all naturally manifest as tensors. These data structures preserve multidimensional relationships through distinct mode axes capturing wavelength, spatial coordinates, temporal frames, viewpoints, or sensor modalities. The exponential growth of such data has intensified the demand for learning models capable of compressing, mining, and analyzing high-order tensors.

\begin{figure}[!h]
  \centering
  \begin{subfigure}{\linewidth}
    \centering
    \includegraphics[width=\linewidth]{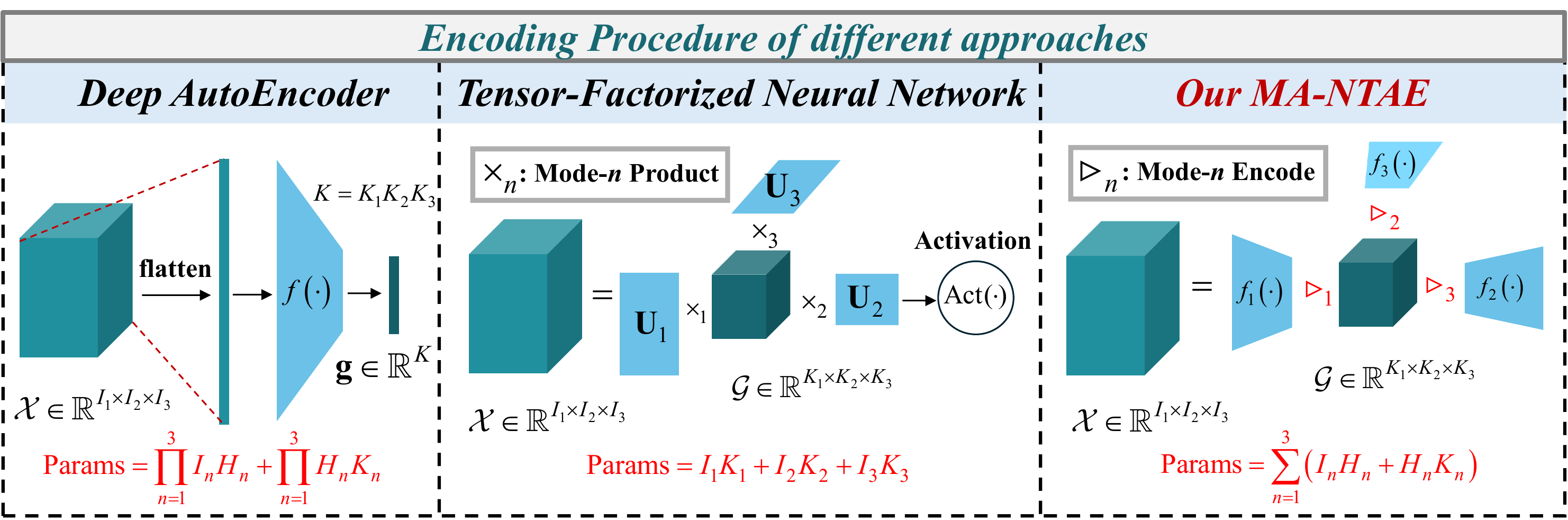}%
    \caption{Foundation of DAE, TFNN, and our proposed approach}
    \label{Comparison_flowchart}
  \end{subfigure}

  \begin{subfigure}{0.49\linewidth}
    \centering
    \includegraphics[width=\linewidth]{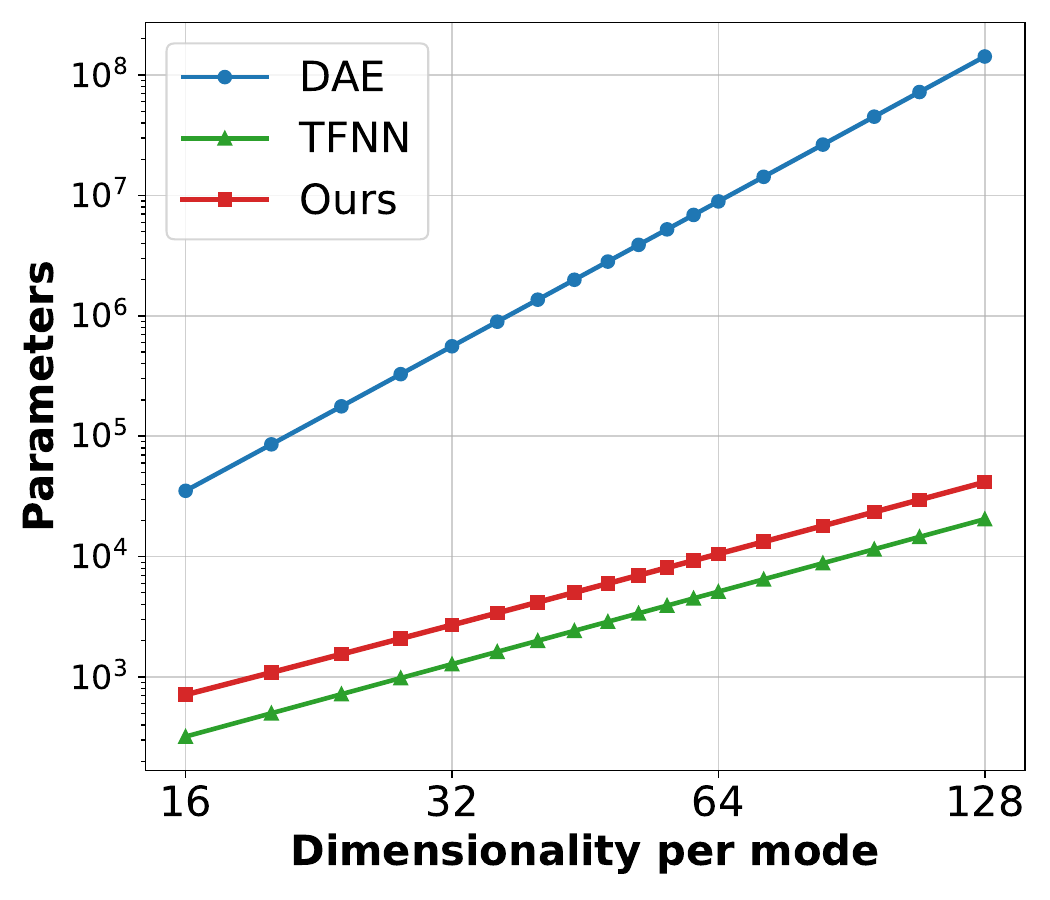}%
    \caption{Parameter growth}
    \label{parameter growth}
  \end{subfigure}\hfill
  \begin{subfigure}{0.49\linewidth}
    \centering
    \includegraphics[width=\linewidth]{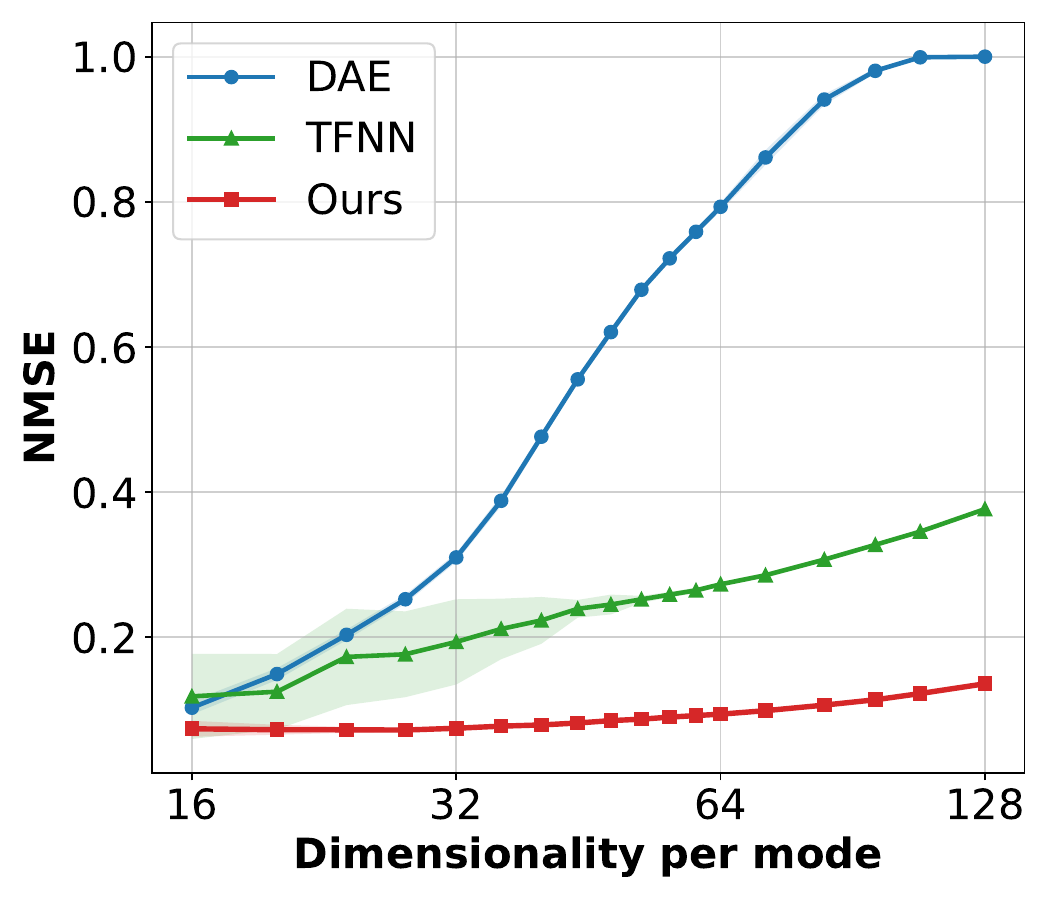}%
    \caption{NMSE vs. Dims}
    \label{NMSE_vs_dim}
  \end{subfigure}

  \caption{Graphical abstract of our innovations and advantages over DAE and TFNN. Our MA-NTAE directly models the non-linear interactions between different modes. (b) and (c) are the results in third-order tensor scenarios (See Synthetic Experiment for details).}
  \label{Graphical abstract}
\end{figure}

Modern deep autoencoders (DAE) based on Multi-layer perceptions (MLPs) \cite{hinton_reducing_2006}, including variants like Variational AEs \cite{VAE} and Adversarial AEs \cite{AAE}, remain dominant in unsupervised representation learning \cite{hu_deep_2025, lin_mhcn_2023}. However, they suffer from two critical limitations when processing tensor-form data: i) \textbf{Mode-agnostic compression}: Flattening operations discard mode-specific statistical dependencies (e.g., temporal correlations versus spatial correlations), which leads to an optimization disaster in recovering structural information; ii) \textbf{Exponential parameter growth}: For an $N^{th}$-order tensor, a fully connected layer mapping flattened input to latent code requires parameters scaling with the multiplication of all input dimension sizes (See the third-order case in Figure \ref{Comparison_flowchart}). This leads to a compromise in the input-data dimensionality among researches \cite{zhu_multiview_2024, wang_adversarial_2023}, where models are also forced to reduce hidden and latent dimensionality to ensure stable convergence.

\subsection{Classical Tucker decomposition revisited}
A naive yet elegant remedy to overcome the curse of dimensionality is offered by the classical multi-linear algebra in \textbf{Tucker decomposition} \cite{tucker1966some}, which factorizes a tensor $\mathcal{X}$ into a core tensor $\mathcal{G}$ and factor matrices $\{\mathbf{U}^{(n)}\}_{n=1}^N$, achieving \emph{linear} parameter growth in tensor order $N$ and \emph{proportional} growth in mode dimensions. Through \textbf{unfold-encode-fold}, the structural information is naturally introduced and integrated into the low-rank approximation for tensor data. During the last decade, researchers have made an effort to utilize Tucker's principle and present tensor autoencoder networks \cite{liu_deep_2022, chien_tensor-factorized_2018, luo_low-rank_2024}. Among them, \citet{chien_tensor-factorized_2018} successfully construct a common Tensor-factorized Neural Network (TFNN) to perform non-linear feature extraction (See Figure \ref{Comparison_flowchart}). However, these approaches are inherently based on linear tensor decomposition frameworks, where neural networks primarily serve to learn the factor matrices for decomposing input data—whether raw inputs or feature tensors extracted by backbone networks. Although these methods introduce non-linear transformations by applying activation functions to the core tensor, they fail to effectively model the non-linear interactions between different modes, ultimately limiting the model's ability to learn complex cross-modal dependencies in the data.

\subsection{Our approach: A Non-linear Tucker Framework}

Inspired by Tucker decomposition and existing tensor networks, we propose \textbf{Mode-Aware Non-linear Tucker Autoencoder (MA-NTAE)}, an intuitive yet effective tensor neural network architecture. A foundation comparison of existing and our approaches is shown in Figure \ref{Comparison_flowchart}. The overall framework of our approach is illustrated in Figure \ref{overall framework}, which embodies three fundamental innovations:

\begin{enumerate}
  \item \textbf{Mode-Aware Non-linear Encoding.}  
        MA-NTAE replaces the global flattening operation in conventional autoencoders by extending Tucker decomposition through a recursively applied \emph{Pick--Unfold--Encode--Fold} strategy. This approach effectively models interactions within individual modes while propagating learned representations across different modes to further explore inter-modal relationships.
  
  \item \textbf{Implicit Structural Priors.}  
        Each time of mode-aware encoding exposes mode-wise covariance structures, where the encoder learns \emph{non-linear Tucker factors} and the folded latent core $\mathcal{X}^{(k)}$ emulates \emph{dynamically optimized core tensor}. By incorporating tensor-structured priors, the proposed method narrows the parameter optimization space, enabling faster and more stable deep mining of tensor data compared to DAEs.
        
  \item \textbf{Low Computational Complexity.}
        MA-NTAE achieves scalable computational complexity that grows linearly with tensor order and proportionally with mode dimensions, while maintaining parameter efficiency - using substantially fewer parameters than DAE and only slightly more than TFNN.
\end{enumerate}

Our main contributions are:
\begin{itemize}
  \item We propose a non-linear Tucker-driven framework that unifies classical tensor factorization with modern autoencoding and allows flexible mode-aware operations in tensor-based unsupervised learning.
  \item We offer a simple yet effective principle---Pick-and-Unfold to handle the curse of dimensionality in higher-order tensor scenarios. 
  \item We provide extensive empirical evidence on synthetic and real tensors demonstrating superior tensor data representation in unsupervised tasks, with advantages that amplify as data dimensionality grows.
\end{itemize}

\section{Related Work}
\noindent \textbf{Notations}. Tensors are denoted by bold calligraphic letters (\(\mathcal X\)), matrices by bold capitals (\(\mathbf X\)), and vectors by bold lower‑case letters (\(\mathbf x\)). $\mathbf{X}^{(n)}\in\mathbb{R}^{I_n \times \prod_{k\neq n}^{N}I_k}$ denotes the mode-$n$ unfolding of $\mathcal{X} \in \mathbb{R}^{I_1 \times \cdots \times I_N}$. 

\noindent \textbf{Deep Autoencoders}. Deep Autoencoders (DAEs) have evolved significantly since their inception as linear dimensionality reducers \cite{bourlard1988auto}. Modern variants includes regularized AEs \cite{vincent2010stacked, rifai2011contractive}, probabilistic AEs \cite{kingma2013auto,makhzani2015adversarial}, and Convolutional AEs \cite{masci2011stacked}. Despite these advances, all flatten high-order tensors into vectors—destroying multi-linear structure and inducing $\mathcal{O}(\prod_{n=1}^N I_n)$ parameter scaling. Our work fundamentally differs by operating natively on tensor manifolds through recursive mode-wise processing.

\noindent \textbf{Tucker Decomposition}. 
Tensor decomposition techniques extract latent structures from high-order data through multi-linear algebraic formulations \cite{kolda2009tensor}. Tucker decomposition \cite{tucker1966some} represents $\mathcal{X}$ as a core tensor $\mathcal{G} \in \mathbb{R}^{K_1 \times \cdots \times K_N}$ multiplied by orthogonal factor matrices $\mathbf{U}_{n} \in \mathbb{R}^{I_n \times K_n}$ along each mode: 
\begin{equation}
  \mathcal{X} \approx \mathcal{G} \times_1 \mathbf{U}_{1} \times_2 \cdots \times_N \mathbf{U}_{N},  
\end{equation}
where $\mathcal{G}\times_n\mathbf{U}_n:=\mathbf{U}_n\mathbf{G}^{(n)}$ is the mode-$n$ product. The multi-linear rank $(K_1, \dots, K_N)$ in Tucker's allows mode-specific compression. Applications based on Tucker decomposition span multiple domains, including image compression \cite{ballester-ripoll_tthresh_2020}, signal processing \cite{haardt_higher-order_2008}, and pattern recognition \cite{hua-chun_tan_expression-independent_2008}. However, the multi-linear operations employed in Tucker decomposition inherently limit its broader application in modern complex downstream tasks.

\noindent \textbf{Tensor-based Neural Network}. Recent advances in \emph{tensor neural networks} (TNNs) show that combining multi-linear algebra with deep learning produces compact, structure‑aware models. 
\citet{chien_tensor-factorized_2018} replace every dense layer with a Tucker factorization followed by an activation function to form a non-linear approximation, preserving mode‑wise correlations while sharply reducing parameters. \citet{ju_tensorizing_2019} leverages tensor train decomposition within a Restricted Boltzmann Machine (RBM) framework to enable non-linear tensor factorization via probabilistic training, improving high-dimensional data modeling. \citet{hyder_compressive_2023} combines tensor ring factorization with a deterministic autoencoder to impose low-rank structural constraints on the latent space, leveraging dataset articulations for improved compressive sensing tasks like denoising and inpainting. 
\citet{zhao_tensorized_2024} tensorizes multi‑view low‑rank approximations so that inter‑view and intra‑class structures are learned jointly, boosting robust hand‑print recognition. Although the above studies employ different tensor decomposition methods and utilize activation functions to model non-linear relationships, their tensor decomposition processes remain fundamentally rooted in linear operations, incapable of achieving a fully non-linear decomposition of tensors that integrates non-linear relationships across modes.

Building on this line, we propose a mode‑aware tensor autoencoder that performs \emph{Pick-Unfold–Encode–Fold} operations, realizing a flexible \emph{non‑linear Tucker compression} with enhanced ability to capture complex non-linear dependencies.

\section{Methodology}
In this section, we formalize the proposed \emph{Mode‑aware Non‑linear Tucker Autoencoder} (MA-NTAE) and detail its optimization. 

\noindent\textbf{Fundamental problem}. The fundamental challenge we address involves developing an efficient tensor compression framework for high-order data representations. Given an $N$-th order tensor $\mathcal X\in\mathbb R^{I_1\times I_2\times\cdots\times I_N}$ ($N\ge 3$), our objective is to learn a non-linear mapping $\mathcal X \rightarrow \mathcal G \in \mathbb{R}^{K_1\times\cdots\times K_N} (K_n<I_n)$  that preserves the intrinsic cross-mode structure while achieving dimensionality reduction. The traditional Tucker decomposition achieves multilinear mapping and reconstruction through a series of mode-specific linear encoders and decoders. Our proposed framework extends this concept to multi-non-linear scenarios by replacing the factor matrices with non-linear mappings:

\begin{equation}
\begin{aligned}
&\mathcal G= \mathcal{X}  \triangleright_1 f_{1} \triangleright_2 \cdots \triangleright_N f_N,\\
&\hat{\mathcal{X}} = \mathcal G \triangleright_N g_N \triangleright_{N-1} \cdots \triangleright_1 g_1,
\end{aligned}
\label{eq:encoderdecoder}
\end{equation}
where $\mathcal X \triangleright_n f_n:=\text{fold}_n\left(f_n\left(\text{unfold}_n(\mathcal X)\right)\right)$, and $f_{n}$ and $g_{n}$ are the mode-specific encoder and decoder sequences, respectively. 

\begin{figure*}
    \centering
    \includegraphics[width=1\linewidth]{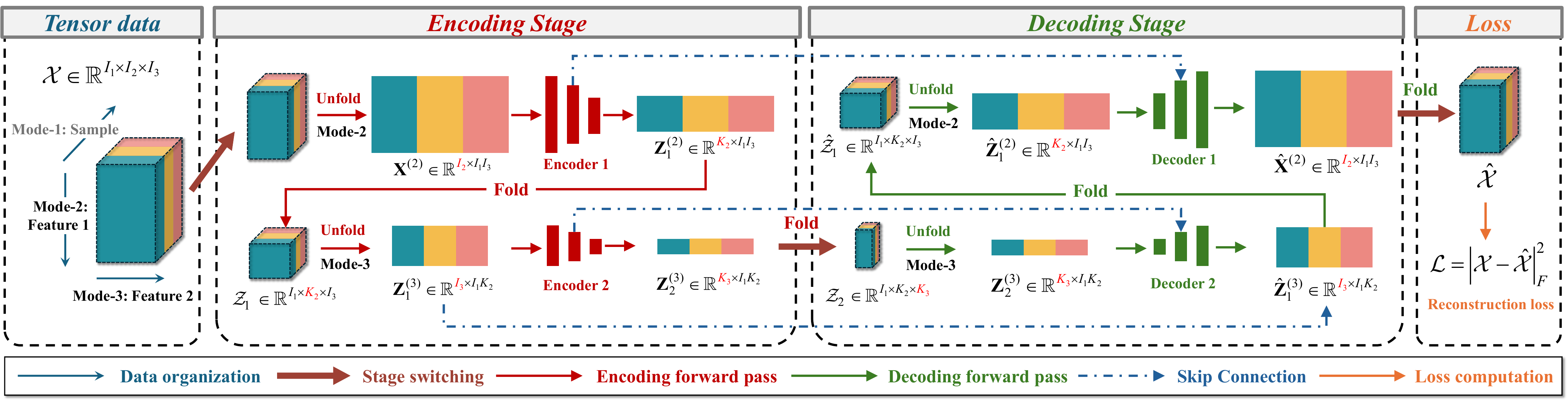}
    \caption{Overall framework of our approach in third-order tensor scenarios. For a batch of tensor data (where each frontal slice represents one sample), we sequentially perform mode-$n$ Unfold--Encode--Fold procedure for each mode, progressively reducing dimensionality across modes. The decoding process follows the reverse mode order to reconstruct data matching the original input dimensions, after which we compute the reconstruction loss. To ensure convergence stability, skip connections are incorporated between corresponding encoder-decoder pairs, leveraging residual learning principles to enhance the network's capacity for modeling high-order tensor data.}
    \label{overall framework}
\end{figure*}

\noindent\textbf{Overview}. Our formulation differs fundamentally from conventional autoencoders that employ vectorization, as MA-NTAE  maintains the tensor organization throughout the transformation process. Figure \ref{overall framework} provides an overview of our approach. The model optimizes the reconstruction $\hat{\mathcal X}=g_\phi(f_\theta(\mathcal X))$ by minimizing reconstruction error (the same as in DAEs), while enforcing compactness in the latent representation $\mathcal G\in\mathbb R^{K_1\times\cdots\times K_N}$.

\subsection{Core Architecture}
\noindent\textbf{Pick--Unfold--Encode--Fold Recursion}. The compression mechanism employs a recursive Pick-Unfold-Encode-Fold procedure that selectively processes individual tensor modes. For an ordered set of target modes $\mathcal S=\{s_1,\dots,s_L\}\subseteq\{1,\dots,N\}$, each compression stage $\ell\in\{1,...,L\}$ executes three key operations:

\begin{enumerate}
\item \textbf{Mode-specific Unfolding}: The current latent tensor $\mathcal Z_{\ell-1} \in \mathbb{R}^{d_i \times \cdots \times d_N}$ with
\begin{equation}
    d_i=
    \begin{cases} 
K_i & i>s_{\ell} \\
I_i & \text{otherwise}
\end{cases}
\end{equation}
undergoes mode-$s_\ell$ unfolding to produce matrix $\mathbf Z_{\ell-1}^{(s_\ell)}\in\mathbb R^{I_{s_\ell}\times J}$ where $J=\prod_{n\neq s_\ell} d_n$. This operation preserves inter-modal correlations while exposing the target mode's features.
\item \textbf{Non-linear Projection}: A dedicated multilayer perceptron processes the unfolded representation:
\begin{equation}
\begin{aligned}
    \mathbf Z_{\ell(s_\ell)} &= \text{MLP}_{\theta_\ell}(\mathbf Z_{\ell-1}^{(s_\ell)})\\ &= \text{FC}_{K_{s_\ell}}(\text{ReLU}(\text{FC}_{H_{s_\ell}}(\mathbf Z_{\ell-1}^{(s_\ell)})))
\end{aligned}
\end{equation}

where $\text{FC}$ refers to Fully Connected layer, and the hidden dimension $H_{s_\ell}$ controls the transformation capacity.

\item \textbf{Structural Reorganization}: The compressed mode is folded back into tensor form $\mathcal Z_{\ell}\in\mathbb R^{K_{s_\ell}\times I_1\times\cdots\widehat{I_{s_\ell}}\cdots\times I_N}$, maintaining proper mode ordering through permutation.
\end{enumerate}

The dimensionality of the tensor progressively decreases with each mode-specific mapping:
\begin{equation}
    \mathcal X \overset{f_1}{\longrightarrow}Z_1\overset{f_2}{\longrightarrow} Z_2 \rightarrow \cdots \overset{f_L}{\longrightarrow}\mathcal Z_L=\mathcal G 
\end{equation}
After $L$ recursive stages, the process yields a compact latent core $\mathcal G=\mathcal Z_L\in\mathbb R^{K_1\times\cdots\times K_N}$. 

\noindent\textbf{Reverse: Pick--Unfold–Decode–Fold Recursion}. The decoder mirrors the encoding procedure in reverse order, employing distinct weights $\phi_\ell$ for each mode's reconstruction network. Correspondingly, the dimensionality of the tensor progressively increases with each mode-specific mapping:
\begin{equation}
    \mathcal G \overset{g_L}{\longrightarrow}\hat{\mathcal{Z}}_{L-1}\overset{g_{L-1}}{\longrightarrow} \hat{\mathcal{Z}}_{L-2} \rightarrow \cdots \overset{g_{1}}{\longrightarrow} \hat{\mathcal{X}}
\end{equation}This architecture generalizes Tucker decomposition by introducing learnable non-linear projections at each factorization step.

\noindent\textbf{Skip Connections for Higher-order Tensor Optimization}. As the order of the input tensor increases, the encoder-decoder chain becomes longer and the network deepens accordingly. To mitigate gradient vanishing and enhance convergence stability for higher-order tensors ($N \geq 4$), we incorporate skip connections between pairwise mode-aware encoder-decoder blocks. The complete algorithmic workflow is presented in \textbf{Algorithm 1}.

\subsection{Loss function and training procedure}
MA-NTAE employs the same loss function as standard DAE, minimizing the reconstruction error:

\begin{equation}
\mathcal L(\theta,\phi) = \frac{1}{B}\sum_{b=1}^{B} |g_\phi(f_\theta(\mathcal X_{b}))-\mathcal X_{b}|_F^2,
\label{eq:mse_loss}
\end{equation}
where $B$ denotes batch size. During training, the proposed model preserves the standard autoencoder training paradigm while operating directly on tensor representations.

\subsection{Computational and Parametric Complexity}
\noindent\textbf{Computational Complexity}. MA‑NTAE performs \emph{mode‑wise} compression: every selected mode
$s_\ell$ is first unfolded, then passes through two linear maps
(\textit{Input}$\rightarrow$\textit{Hidden}$\rightarrow$\textit{Latent}), and is finally folded
back. The exact floating‑point cost for this mode is

\begin{equation}
\begin{aligned}
\mathrm{FLOPs}_{\text{enc}}(s_\ell)
&=\underbrace{\mathcal O\!\bigl(I_{s_\ell}\,D_{-s_\ell}\bigr)}_{\text{unfold}}
 + I_{s_\ell}H_{s_\ell}D_{-s_\ell} \\
 &+ H_{s_\ell}K_{s_\ell}D_{-s_\ell}
 +\underbrace{\mathcal O\!\bigl(K_{s_\ell}\,D_{-s_\ell}\bigr)}_{\text{fold}} \\[2pt]
&\approx D_{-s_\ell}\,H_{s_\ell}\,(I_{s_\ell}+K_{s_\ell}),
\end{aligned}
\label{eq:mode_flops}
\end{equation}

\noindent
where the unfold/fold terms are linear in the element count and therefore
dominated by the two matrix products in most practical settings. Summing~\eqref{eq:mode_flops} over all $N$ modes yields

\begin{equation}
\mathrm{FLOPs}_{\text{enc}}
  = \sum_{s_\ell=1}^{L} H_{s_\ell}D_{-s_\ell}\left(I_{s_\ell} + K_{s_\ell}\right)
  = \mathcal O\!\Bigl(
        L\,\overline{H}_{s_\ell}\,\overline{I}^{\,N}
     \Bigr),
\end{equation}

\noindent
where $\overline{I}$ and $\overline{H}$ are the representative mode and hidden size in the regular
case ($I_n=\overline{I}, H_n=\overline{H}$).  The decoder is symmetric and contributes
the same asymptotic cost. Therefore, in the extreme case where $L=N$, the overall complexity of MA‑NTAE remains \textbf{linear in tensor order $N$} and \textbf{proportional to each mode dimension~$I_n$}. 

\noindent\textbf{Parameter Complexity}. Per compressed mode $s$ the encoder holds two matrices
$H_s\times I_{s}$ and $K_{s}\times H_s$ and the decoder holds their
transposes, so biases aside
\begin{equation}
    \mathrm{Params}(s)=2H_s\bigl(I_{s}+K_{s}\bigr).
\end{equation}
Summing over all modes gives the network size
\begin{equation}
\mathrm{Params}_{\text{MA--NTAE}}=2\sum_{n=1}^{N}H_n(I_n+K_n),
\end{equation}
linear in the tensor order $N$ and in each mode dimension $I_n$. Figure \ref{parameter growth} compares the parameter growth of DAE, TFNN, and our approaches. Our method achieves substantially greater parameter efficiency compared to DAE while maintaining a marginally larger parameter count than TFNN.

\section{Experiments}
We assess theoretical performance on synthetic tensor datasets and validate effectiveness on real-world measurements through compression and clustering experiments. We conduct DAE and TFNN for comparison. We utilize PyTorch \cite{pytorch} to implement our method and an NVIDIA RTX 4090 GPU to run each experiment under Windows 10 operating system. 

\noindent\textbf{Implementary details}. We conduct MA-NTAE with a dimensionality reduction factor $\alpha$ and set $I-\alpha I-\alpha^2I$ per mode-wise encoder (up to mode-$N-1$, not including sample mode). The corresponding decoder layers are set in a reverse fashion. For TFNN, we adapt the structure from \cite{chien_tensor-factorized_2018} and construct a tensor autoencoder that maintains identical layer configurations and tensor dimensionality to MA-NTAE. We conduct DAE with the same number of neurons ($I^{N-1}-(\alpha I)^{N-1}-(\alpha^2 I)^{N-1}-(\alpha I)^{N-1}-I^{N-1}$). \textbf{All comparative models optimize the MSE as the loss function}, while normalized MSE (NMSE) is utilized for evaluation. We consistently employ the Rectified Linear Unit (ReLU) function as the activation function for all methods.

\subsection{Synthetic Experiment}
\noindent\textbf{Data formulation}. To evaluate MA-NTAE's feasibility and robustness, we synthesize $N$th-order tensors of shape $(B,I,\ldots,I)$, where $B=512$ is the batch size and $I$ tests spatial resolutions. The Tucker core maintains shape $512\times0.25I\times\cdots\times0.25I$ for consistent compression. For each sample, we generate $N-1$ orthonormal factor matrices $\mathbf{U}^{(n)}\in\mathbb{R}^{I\times0.25I}$ ($n=2,\ldots,N$), perturb them with Gaussian noise ($\sigma_U=0.05$) to obtain $\tilde{\mathbf{U}}^{(n)}$, then construct clean tensors via:

\begin{equation}
\mathcal{X}^{(b)}_{\mathrm{clean}} = \mathcal{G}^{(b)} \times_{2}\tilde{\mathbf{U}}^{(2)} \times_{3}\cdots\times_{N}\tilde{\mathbf{U}}^{(N)},\quad \mathcal{G}^{(b)}\sim\mathcal{N}(0,1).
\end{equation}
We then add 30dB Gaussian noise to create $\mathcal{X}^{(b)}_{\mathrm{noisy}}=\mathcal{X}^{(b)}_{\mathrm{clean}}+\Delta$. This setup generalizes the evaluation to arbitrary tensor orders while preserving the original noise and compression constraints. We compute MSE between $X_{\mathrm{noisy}}$ and $\hat{X}_{\mathrm{noisy}}$ as the loss function and NMSE between $\hat{X}_{\mathrm{noisy}}$ and $X_{\mathrm{clean}}$ for evalution. For each synthetic tensor, we allocate $80\%$ of noisy samples for training and $20\%$ for testing (clean tensors split identically). For each setting of $I$, we repeat the experiment $30$ times and average the results to avoid statistical bias. 

\begin{figure}[tb]
  \centering
  \begin{subfigure}{0.48\linewidth}
      \includegraphics[width=\linewidth]{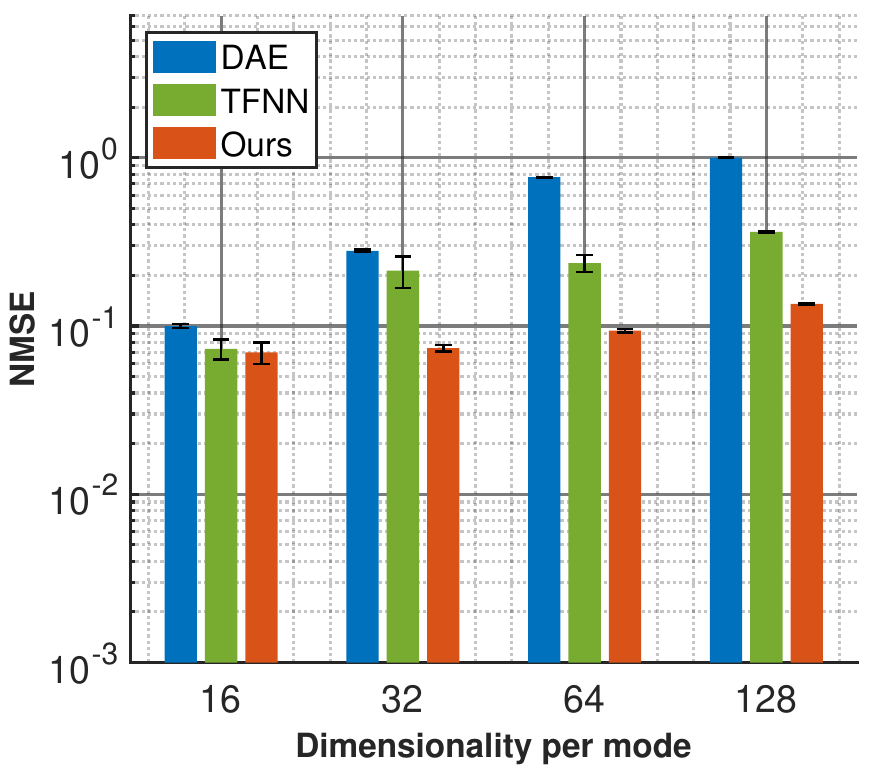}%
    \caption{NMSE (no permuted)}
    \label{NMSE_test_0}
  \end{subfigure}\hfill
  \begin{subfigure}{0.48\linewidth}
    \centering
    \includegraphics[width=\linewidth]{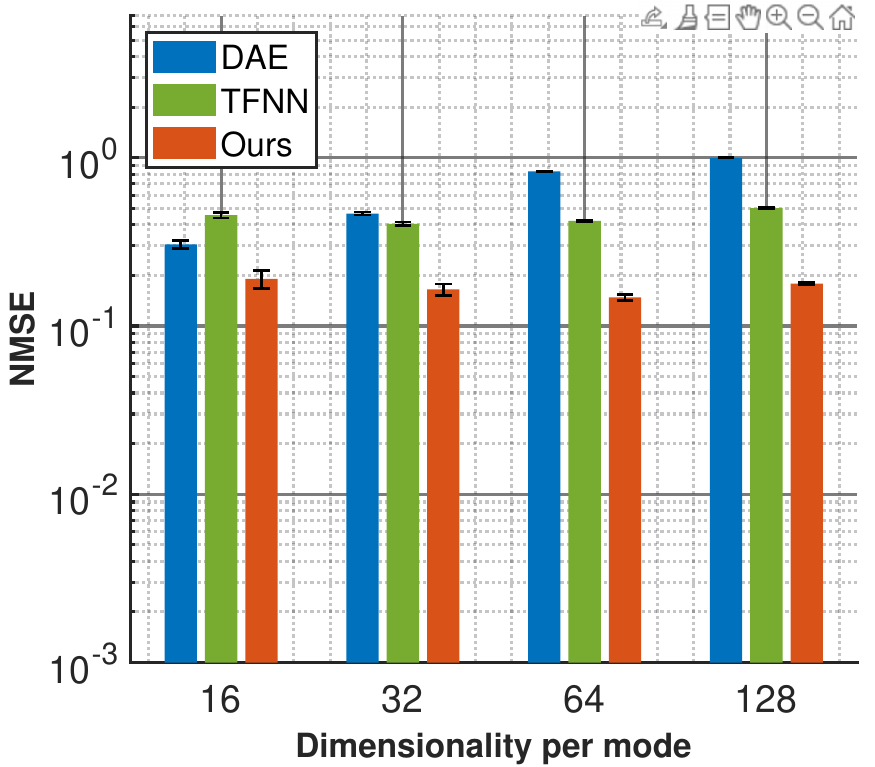}%
    \caption{NMSE ($10\%$ permuted)}
    \label{NMSE_test_10}
  \end{subfigure}\\
    \begin{subfigure}{0.48\linewidth}
      \includegraphics[width=\linewidth]{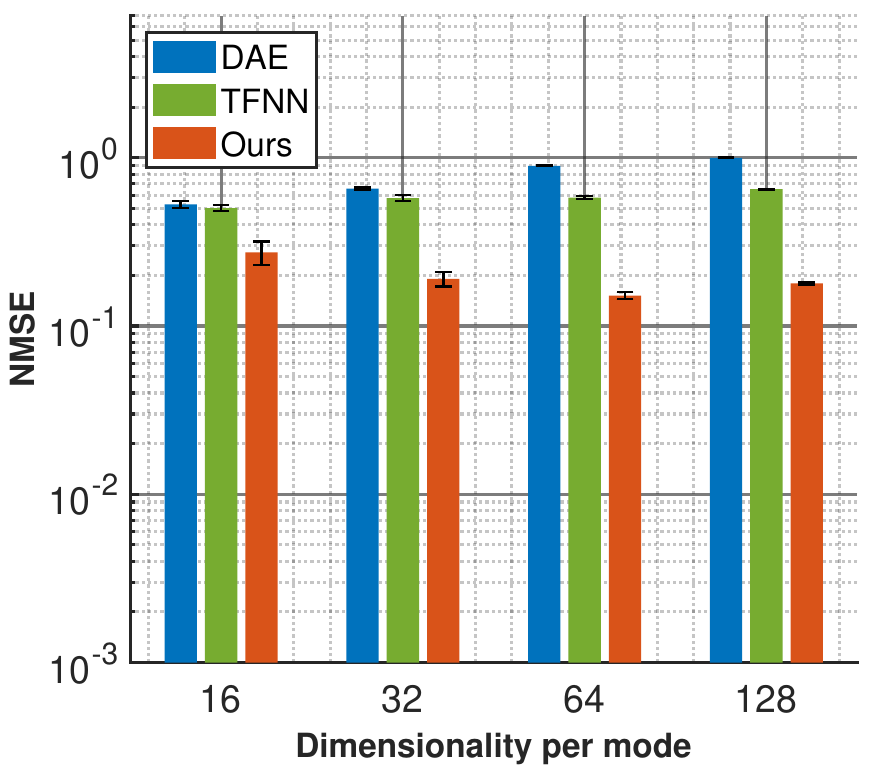}%
    \caption{NMSE ($20\%$ permuted)}
    \label{NMSE_test_20}
  \end{subfigure}\hfill
  \begin{subfigure}{0.48\linewidth}
    \centering
    \includegraphics[width=\linewidth]{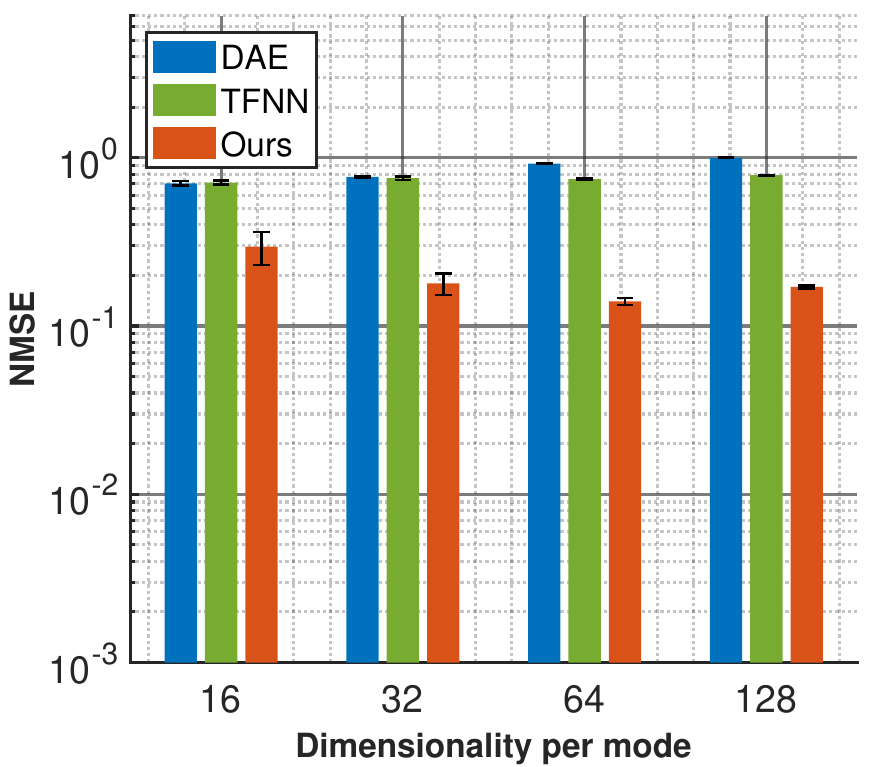}%
    \caption{NMSE ($30\%$ permuted)}
    \label{NMSE_test_30}
  \end{subfigure} 

  \caption{NMSE on the test set of third-order synthetic tensor data with random mode permutation. We randomly select a subset of samples, shuffle their mode orders, and evenly distribute them between the training and test sets.}
  \label{NMSE_random_mode_permutation}
\end{figure}

\noindent\textbf{Results}. 
Figure \ref{NMSE_vs_dim} and Table \ref{tab:nmse_time_ae_tae} demonstrates our method's superior noise robustness and low computational cost on tensor structure recovering. The performance gap between ours and comparative methods widens with dimensionality and tensor orders. Figure \ref{NMSE_random_mode_permutation} reveals that mode-shuffled samples degrade performance for all methods, with mode-wise methods (TFNN and our approach) being more sensitive to incorrect ordering. By direct non-linear tensor decomposition, our approach achieves a more stable NMSE growth trend with varying dimensionality and tensor orders while maintaining satisfying training time.

\begin{table*}[tb]
\centering
\caption{NMSE($\pm \text{std}$) and training time (per epoch, seconds) on tensors of different orders. Dimension per mode is set to $20$. }
\label{tab:nmse_time_ae_tae}
\begin{tabular}{ccccccc}
\toprule
\multirow{2}{*}{\textbf{Order}}
& \multicolumn{2}{c}{\textbf{DAE}} & \multicolumn{2}{c}{\textbf{TFNN}}
& \multicolumn{2}{c}{\textbf{Ours}} \\
\cmidrule(lr){2-7} 
& \textbf{NMSE} & \textbf{Time} & \textbf{NMSE} & \textbf{Time} & \textbf{NMSE} & \textbf{Time} \\
\midrule
3 & $0.1467\pm0.0050$ & $0.0094$ & $0.1249\pm0.0520$ & $0.0124$ &$0.0743\pm0.0080$ &$0.0209$\\
4 & $0.6435\pm0.0037$ & $0.0268$ & $0.1517\pm0.0016$ & $0.0186$ &$0.1005\pm0.0187$ &$0.0584$\\
5 & $1.0023\pm0.0006$ & $59.2248$ & $0.2870\pm0.0020$ & $0.4833$ &$0.2440\pm0.0338$ &$0.5296$\\
\bottomrule
\end{tabular}
\end{table*}

\subsection{Experiment on real-world data} 
\subsubsection{Visual Image Compression}

We first carry out a visual image compression experiment on the multi-view object image dataset COIL20 \cite{COIL20}, and two facial datasets--JAFFE for expression analysis\cite{jaffe} and Orlraws10P\footnote{https://jundongl.github.io/scikit-feature/datasets.html\label{orlraws10p}} with pose variations. The real-world datasets we used are detailed in Table \ref{tab:datasets}. All image data were used without any preprocessing except for normalization to the interval of $[0,1]$. We employed a balanced $50-50$ split for training and testing sets to ensure equitable data distribution. All methods are trained for $1000$ epochs.

\begin{figure*}[h]
\centering
\includegraphics[width=\textwidth]{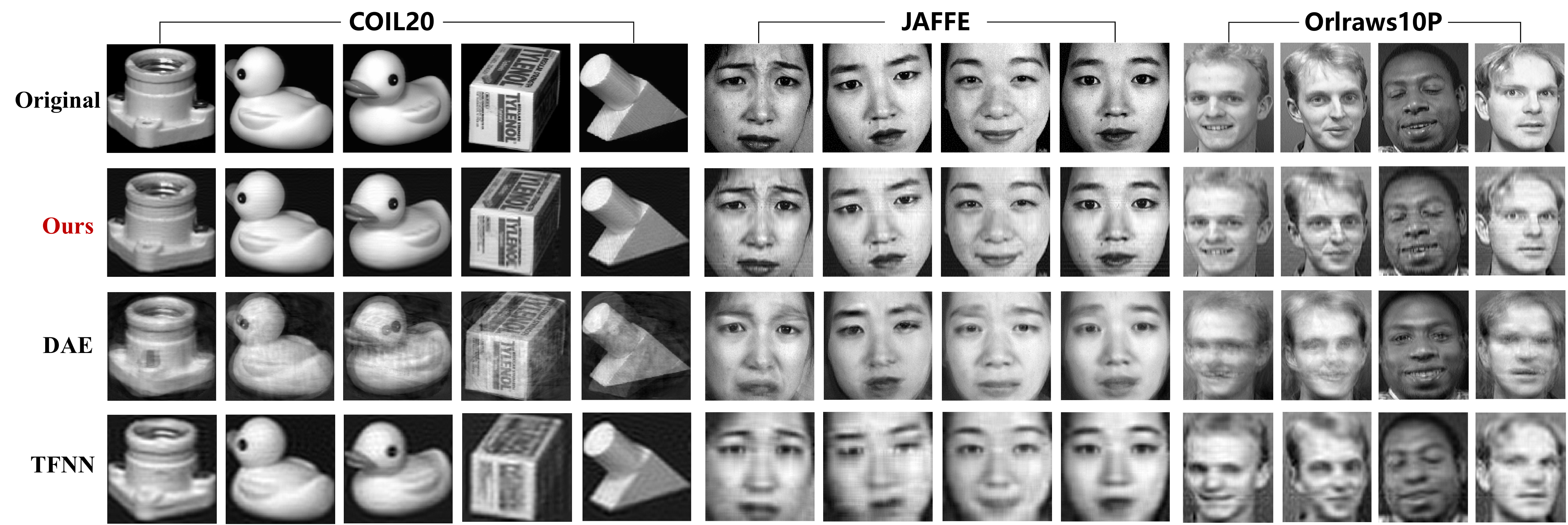}
\caption{The reconstruction results of comparative methods on the test sets of COIL20, JAFFE, and Orlraws10P. The compression ratio is set to $16/1$ by adjusting the dimensionality reduction factor to $0.5$.}  
\label{image_compression}
\end{figure*}

\begin{figure}
    \centering
    \begin{subfigure}{0.5\linewidth}
        \includegraphics[width=\linewidth]{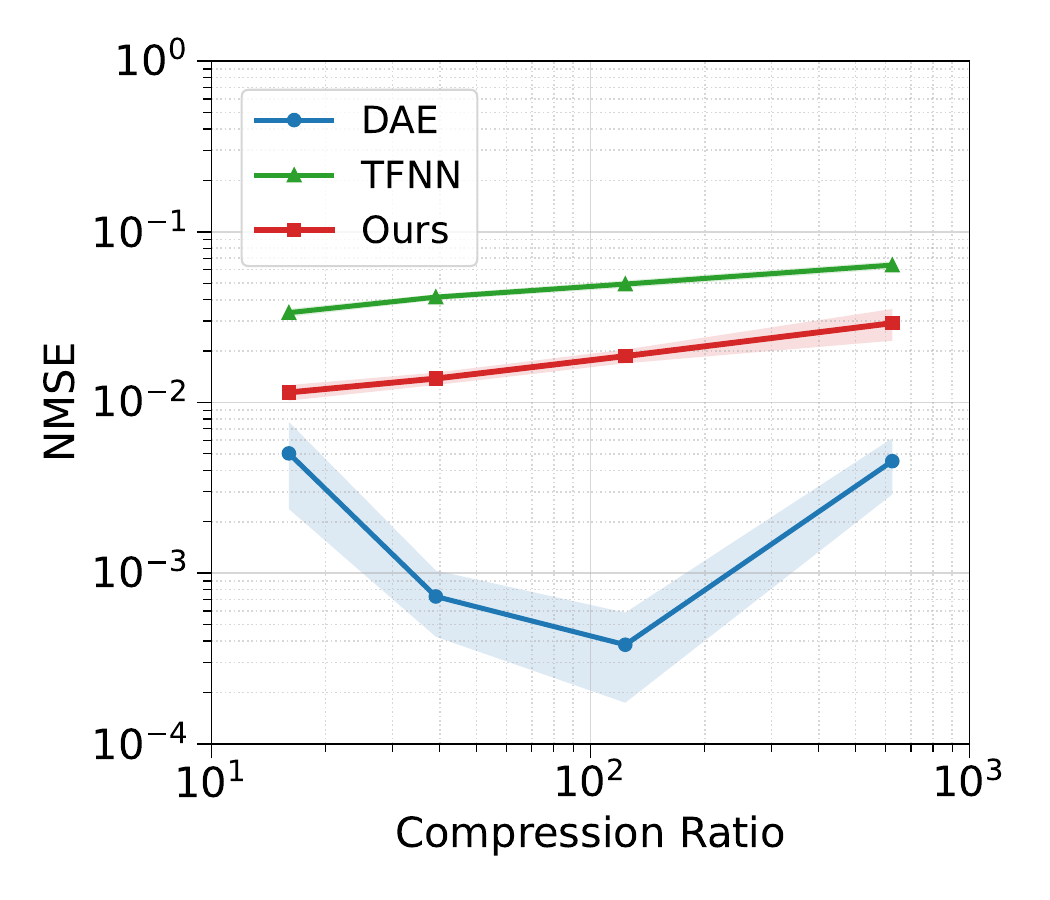}
    \caption{NMSE on the train set}
    \end{subfigure}\hfill
    \begin{subfigure}{0.5\linewidth}
        \includegraphics[width=\linewidth]{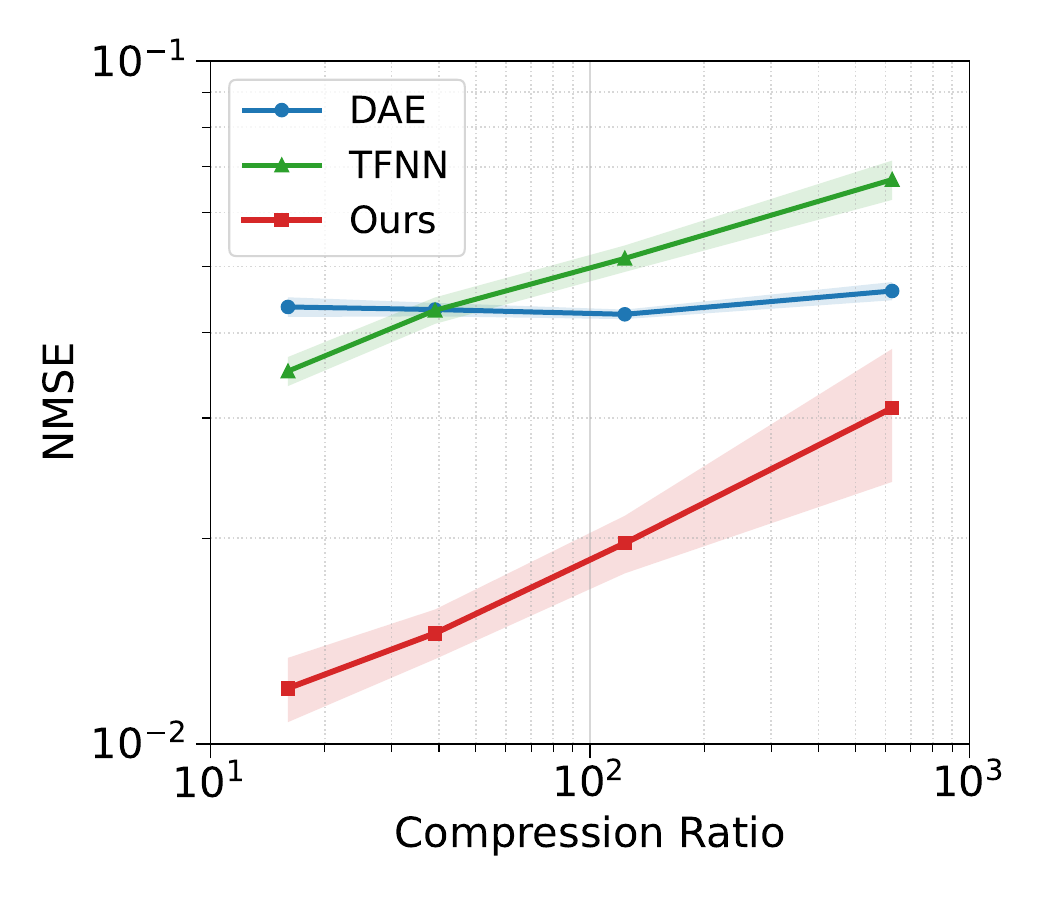}
    \caption{NMSE on the test set}
    \end{subfigure}
    \caption{NMSE vs. Compression Ratio on Orlraws10P. } 
    \label{cr_vs_nmse_loglog}
\end{figure}

\begin{figure}
    \centering
    \begin{subfigure}{0.5\linewidth}
        \includegraphics[width=\linewidth]{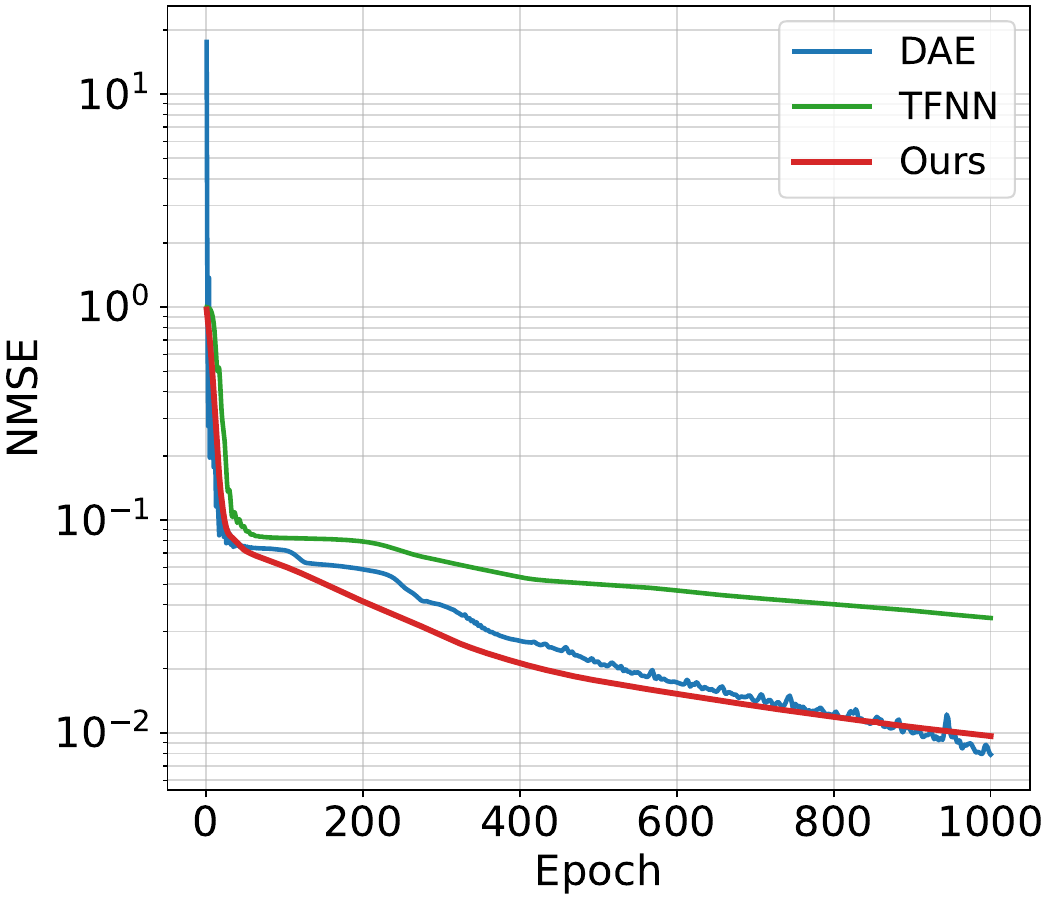}
    \caption{Training loss (normalized)}
    \end{subfigure}\hfill
    \begin{subfigure}{0.5\linewidth}
        \includegraphics[width=\linewidth]{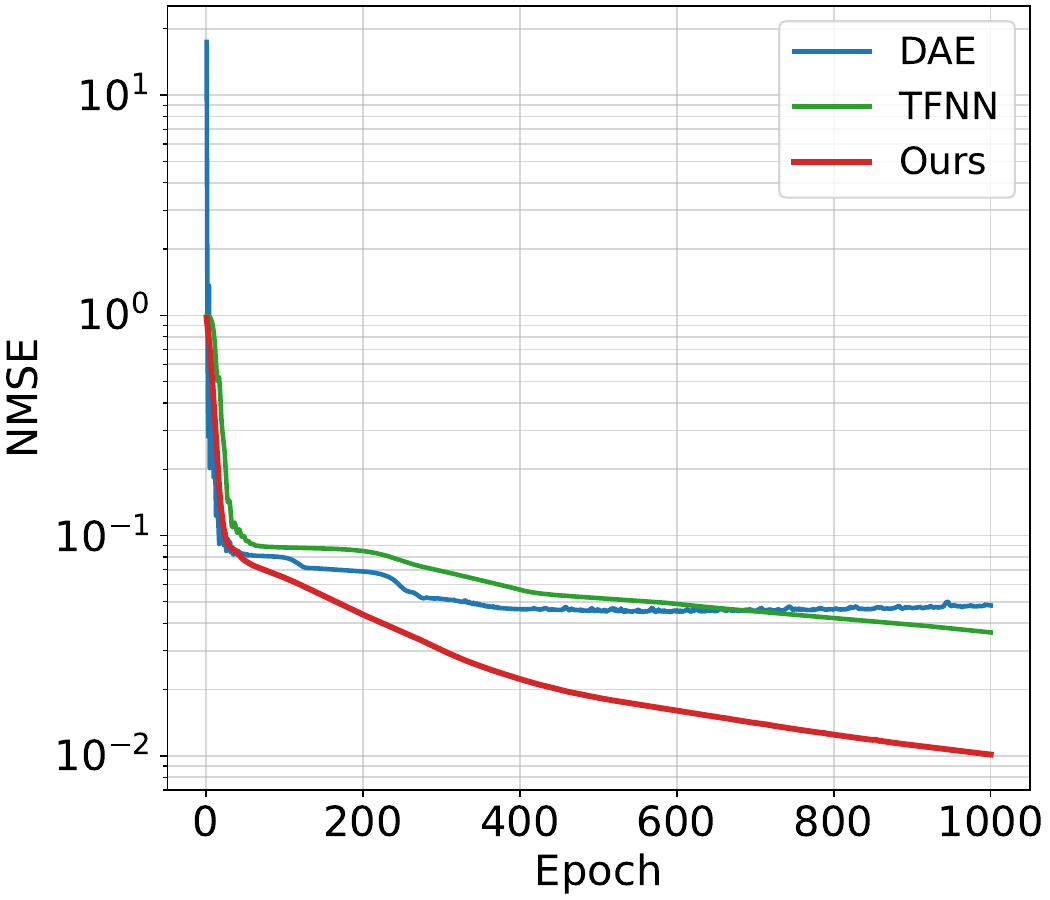}
    \caption{Test loss (normalized)}
    \end{subfigure}
    \caption{Loss curves on the training and test set of Orlraws10P.}
    \label{loss}
\end{figure}

\noindent \textbf{Results}. Figure \ref{image_compression} demonstrates that our approach exhibits significantly better adaptability across varying viewpoints and poses compared to DAE (which suffers from varying degrees of view confusion and target ambiguity across all datasets) and TAE (which obtains blurred images). Further, we vary the compression ratio by setting the dimensionality reduction factor in the range of $[0.5,0.4,0.3,0.2]$, and repeated the experiments $30$ times to obtain the NMSE curves in Figure \ref{cr_vs_nmse_loglog}. While DAE achieves lower reconstruction error on the training set, its performance degrades significantly on the test set compared to MA-NTAE. This explains why DAE erroneously reconstructs some test samples as training images - a clear manifestation of overfitting. By leveraging the tensor structures, By explicitly exploiting the inherent tensor structures, our method achieves (1) superior compression and reconstruction performance, and (2) more stable training convergence (See Figure \ref{loss}) and relatively less training time (See Table \ref{traing time}). The compression experiments preliminarily demonstrate the proposed method's promising application potential for real-world tensor-structured data, particularly in multi-view scenarios.

\begin{table}[tb]
\centering
\caption{Dataset Statistics}
\label{tab:datasets}
\begin{tabular}{cccc}
\toprule
\textbf{Dataset} & \#\textbf{Sample} & \#\textbf{Feature} & \#\textbf{Class} \\
\midrule
COIL20 & 1440 & $128\times128$ & 20 \\
JAFFE & 213 & $128\times128$ & 7 \\
Orlraws10P & 100 & $92\times112$ & 10 \\
PIE & 1166 & $32\times 32$ & 53 \\
\bottomrule
\end{tabular}
\end{table}

\begin{table}[tb]
\centering
\caption{Training time per epoch (seconds) on real-world datasets}
\label{traing time}
\begin{tabular}{cccc}
\toprule
{\textbf{Dataset}}
& \textbf{DAE} & {\textbf{TFNN}}
& {\textbf{Ours}} \\
\midrule
COIL20 & $0.7197$ & $0.0386$ & $0.0690$\\
JAFFE & $0.1217$ & $0.0064$ & $0.0067$\\
Orlraws10P & $0.0267$ & $0.0031$ & $0.0031$\\
Traffic & $1.5083$ & $0.0366$ & $0.0408$ \\
\bottomrule
\end{tabular}
\end{table}

\subsubsection{Video Compression} To validate our algorithm's applicability to higher-order real-world data, we conducted a video compression experiment. We employ a standard benchmark video from MATLAB's built-in dataset \footnote{This video is accessible via a MATLAB command \texttt{trafficVid = VideoReader('traffic.mj2')}}, consisting of 120 grayscale frames with a spatial resolution of 120×160 pixels. This sequence captures typical urban traffic patterns, providing realistic motion characteristics for evaluating temporal compression performance. All methods are trained for $1000$ epochs.

\noindent \textbf{Implementary details}. We partition the sequence into overlapping 3-frame snippets as training samples. We set the dimensionality reduction factor per mode encoder layer to $0.3$ for MA-NTAE, accordingly adjusting DAE and TFNN. We selectively encode only spatial modes to demonstrate our method's mode-aware processing capability.

\begin{figure}
    \centering
    \includegraphics[width=\linewidth]{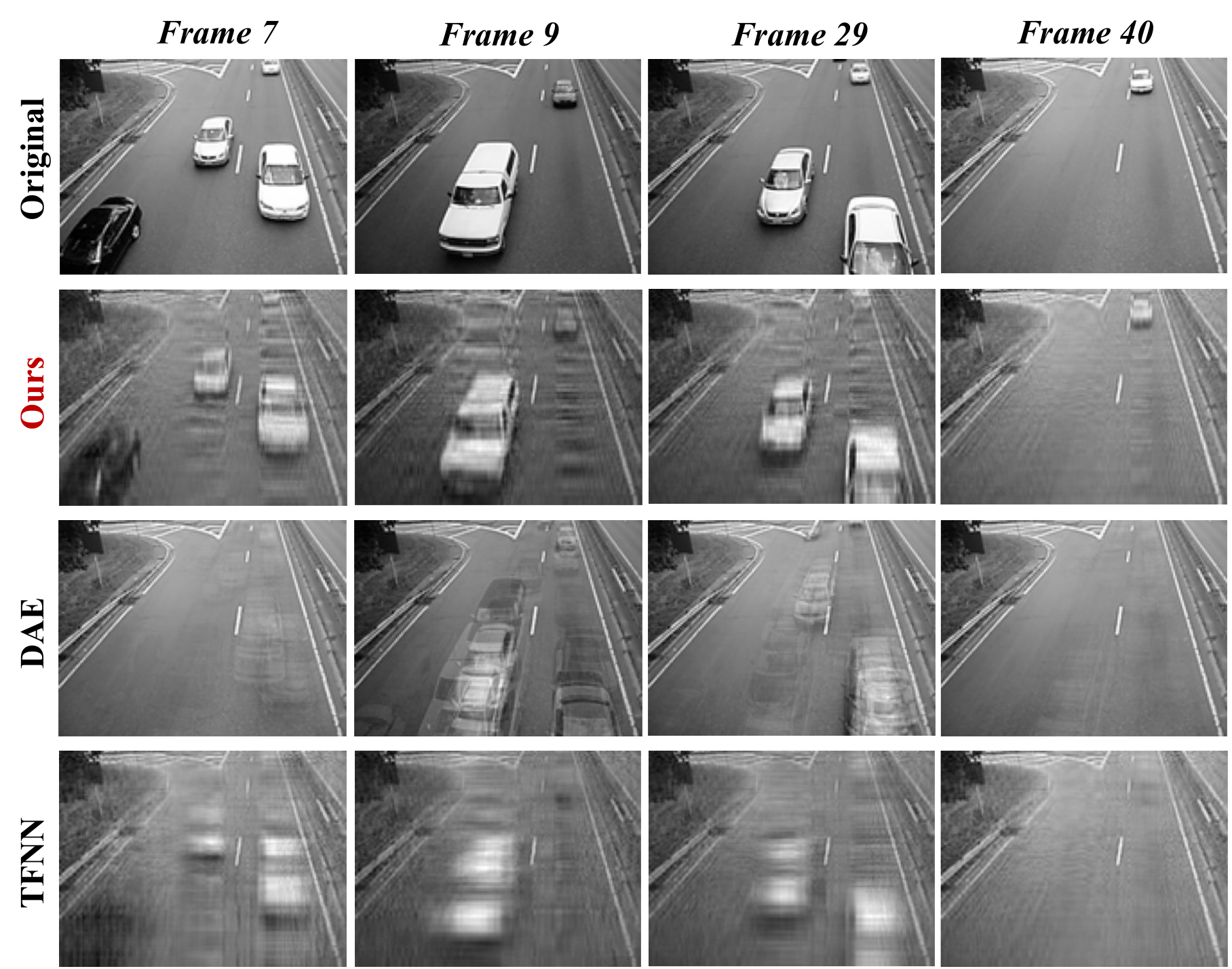}
    \caption{Reconstruction results on video data. We retrieve typical frames containing vehicles in motion for analysis.}
    \label{video_compression}
\end{figure}

\noindent \textbf{Results}. The reconstruction results of representative video frames are shown in Figure \ref{video_compression}. Our method demonstrates superior performance in preserving moving object contours and positional information compared to baseline approaches. Notably, in frame 40, both DAE and TFNN fail to reconstruct the distant vehicle. While DAE achieves the best background detail preservation, it produces significant ghosting artifacts that obscure vehicle positions. TFNN, benefiting from tensor structure utilization, can approximately localize vehicles but generates overly blurred reconstructions due to limited non-linear fitting capacity, resulting in substantial detail loss.

\subsubsection{Visual Image Clustering}

\begin{figure}
    \centering
    \begin{subfigure}{0.5\linewidth}
        \centering
        \includegraphics[width=\linewidth]{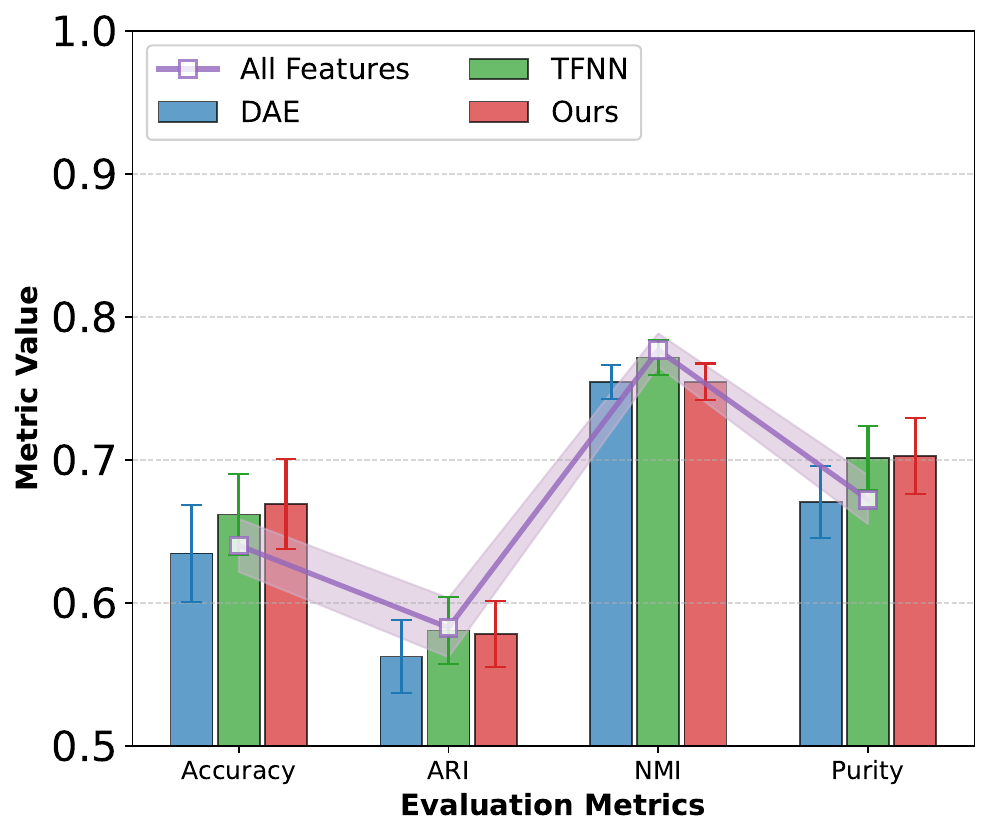}
        \caption{COIL20}
        \label{coil20}
    \end{subfigure}\hfill
    \begin{subfigure}{0.5\linewidth}
        \centering
        \includegraphics[width=\linewidth]{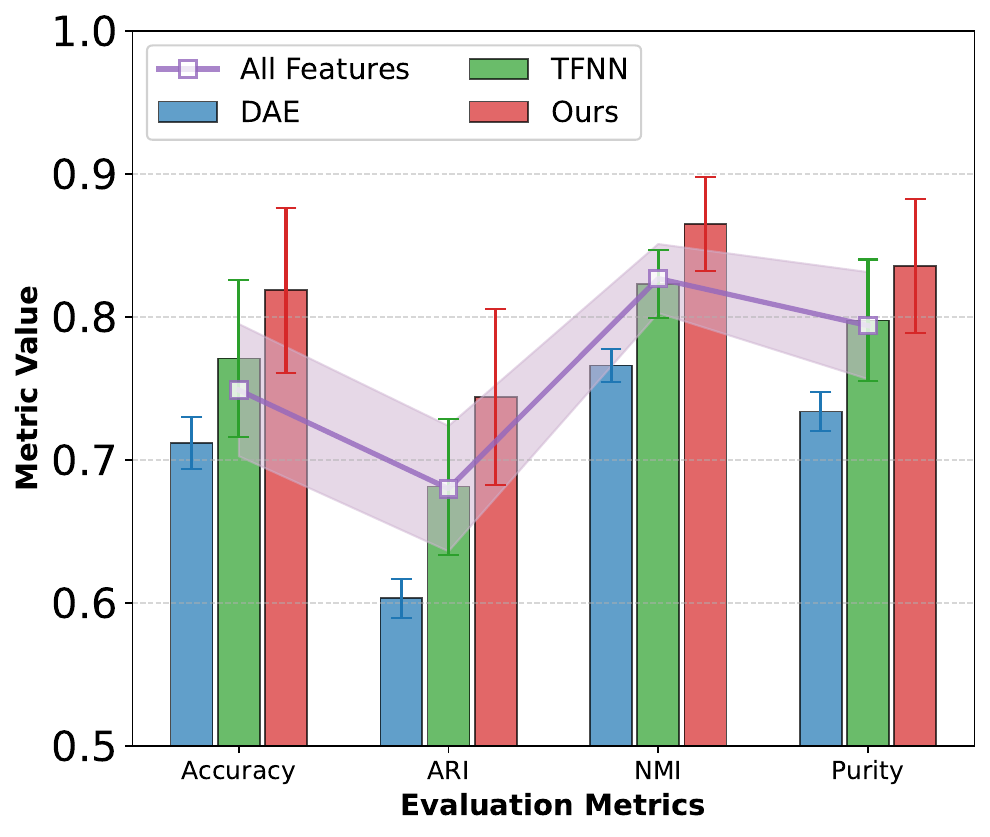}
        \caption{JAFFE}
        \label{jaffe}
    \end{subfigure}

    \vspace{1em} 

    \begin{subfigure}{0.5\linewidth}
        \centering
        \includegraphics[width=\linewidth]{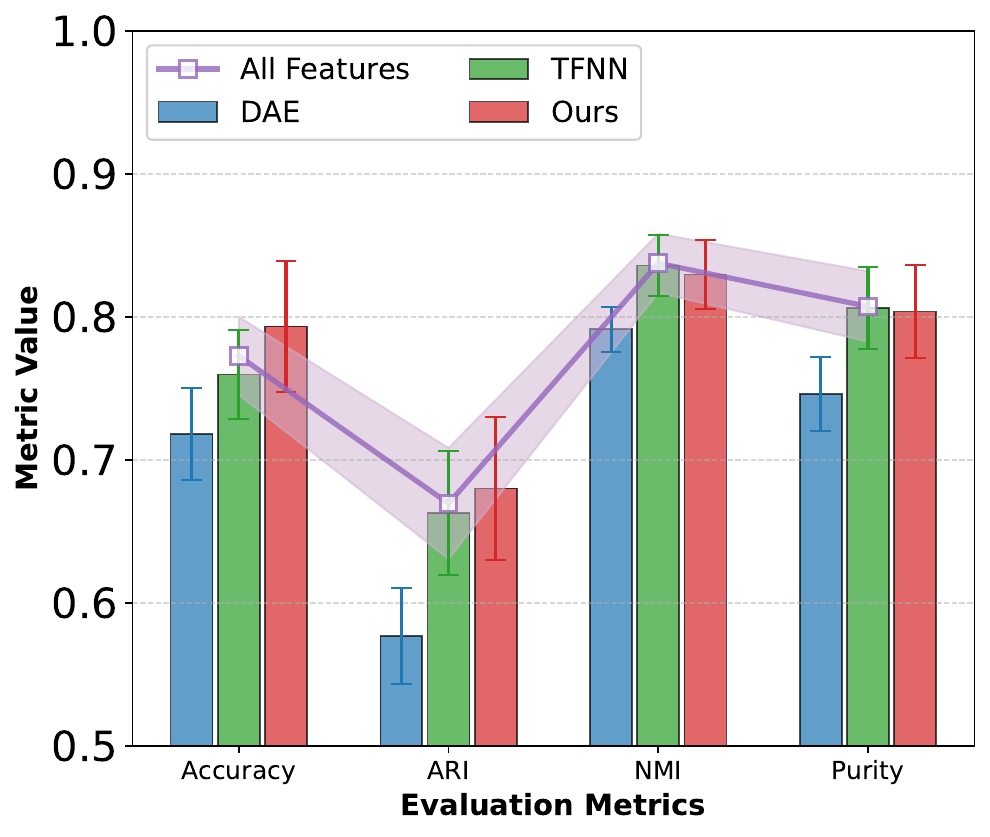}
        \caption{Orlraws10P}
        \label{orl}
    \end{subfigure}\hfill
    \begin{subfigure}{0.5\linewidth}
        \centering
        \includegraphics[width=\linewidth]{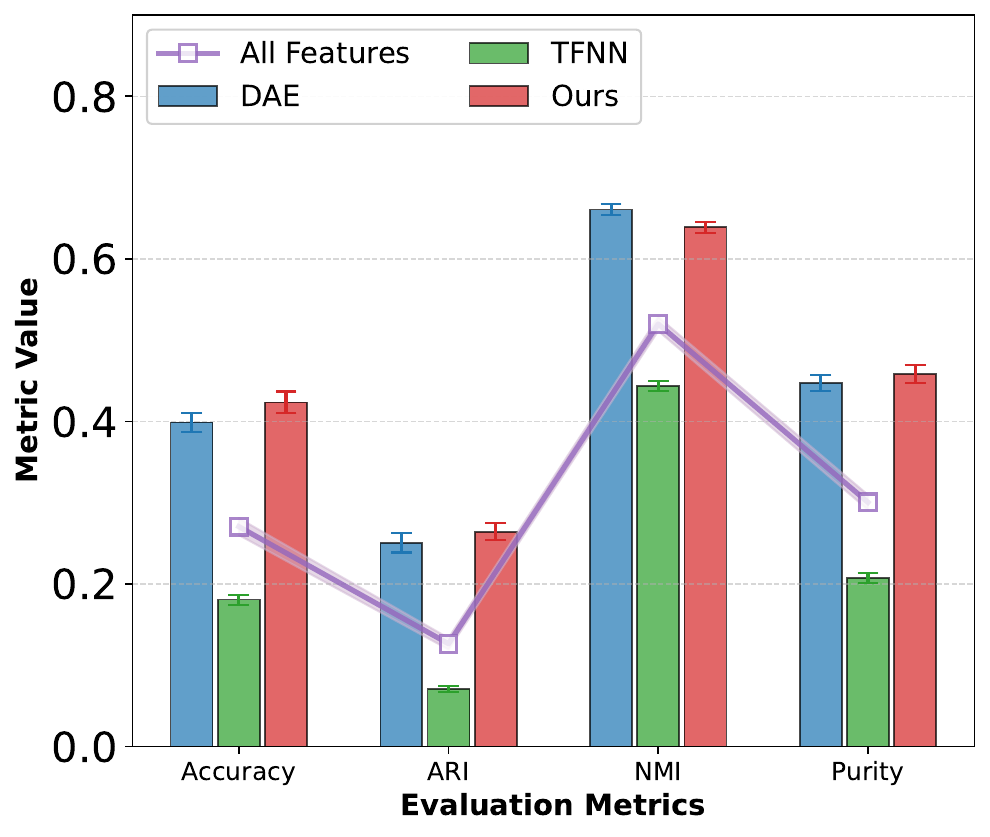}
        \caption{PIE}
        \label{pie}
    \end{subfigure}
    \caption{Clustering results on real-world datasets using \textbf{solely the reconstruction error} as the loss function.}
    \label{clustering}
\end{figure}

In this section, we conduct clustering experiments on COIL \cite{COIL20}, JAFFE \ref{jaffe}, Orlraws10P, and PIE \cite{PIE}. For Orlraws10P and JAFFE, the dimensionality reduction factor is set to $0.5$. For COIL20 and PIE, to avoid excessive reconstruction fitting performance, we adjust the dimensionality reduction factor to $1/3$ and $1/4$, respectively. The minimum number of latent features was set to 25, preventing over-compression. We randomly allocate 80\% of the samples for training. The training epoch is set to $500$ on Orlraws10P and PIE, and $1000$ on COIL20 and JAFFE. After training, all samples are used for clustering tests with K-means. We evaluate the results using clustering metrics: Accuracy, Adjusted Rand index (ARI) \cite{hubert_comparing_1985}, Normalized Mutual Information (NMI) \cite{4309069}, and Purity. The clustering is repeated $30$ times, and the average results are recorded. We use \emph{All Features} as a baseline method, which uses all features to perform clustering.

\noindent \textbf{Results}. Figure \ref{clustering} shows the clustering results. Compared with DAE and TFNN, the proposed method achieves the highest accuracy in clustering tasks across multiple datasets after being trained with the reconstruction loss. Meanwhile, in terms of ARI, NMI, and Purity, it exhibits performance levels that are either superior to or close to those of other encoders. Particularly on the JAFFE dataset, the proposed encoder significantly outperforms DAE, TFNN and the original clustering results in all indicators. Such clustering results are consistent with the fact that our method yields the smallest reconstruction error and the best reconstruction performance on the test set in the reconstruction task, indicating that our method achieves dual advantages: (1) higher computational and training efficiency; (2) the ability to extract unique features of different samples while preserving the sample structure. The k-means clustering experiments preliminarily demonstrate the application potential of the proposed method in the field of feature engineering and downstream tasks.

\section{Conclusion}

In this work, we address the challenges of unsupervised learning on high-order tensor data by proposing the Mode-Aware non-linear Tucker Autoencoder (MA-NTAE), a novel framework that integrates classical Tucker decomposition with modern autoencoding techniques through recursive Pick-Unfold-Encode-Fold operations and enables flexible mode-aware processing of tensor data. Compared to DAE (vector-based) and existing Tucker-based tensor network: TFNN, our approach achieves superior reconstruction accuracy with relatively small parameter sizes and training time across simulated and real-world tensor data of varying orders and dimensions. For multi-view image data, it effectively reconstructs both viewing angles and fine details in test samples. When processing video data, the method demonstrates an enhanced capability to balance motion target localization and contour refinement. Notably, in clustering tasks, it delivers better overall clustering metrics using only the reconstruction error loss function. Future work will explore integrating more DAE-proven variants into our Pick-and-Unfold tensor autoencoder framework to enable broader specialized applications.

\end{document}